# COFAP: A Universal Framework for COFs Adsorption Prediction through Designed Multi-Modal Extraction and Cross-Modal Synergy

Zihan Li[1,2], Mingyang Wan[1,2], Mingyu Gao[4], Xishi Tai[5], Zhongshan Chen[3*], Xiangke Wang[3*] and Feifan Zhang[1*]

**ABSTRACT Covalent organic frameworks (COFs) are promising adsorbents for gas adsorption and separation, while identifying the optimal structures among their vast design space requires efficient high-throughput screening. Conventional machine-learning predictors rely heavily on specific gas-related features. However, these features are time-consuming and limit scalability, leading to inefficiency and labor-intensive processes. Herein, a universal COFs adsorption prediction framework (COFAP) is proposed, which can extract multi-modal structural and chemical features through deep learning, and fuse these complementary features via cross-modal attention mechanism. Without relying on explicit gas-specific thermodynamic descriptors, COFAP achieves state-of-the-art prediction performance on the hypoCOFs dataset under the conditions investigated in this study, outperforming existing approaches. Based on COFAP, we also found that high-performing COFs for gas separation concentrate within a narrow range of pore size and surface area. A weight-adjustable prioritization scheme is also developed to enable flexible, application-specific ranking of candidate COFs for researchers. Superior efficiency and accuracy render COFAP directly deployable in crystalline porous materials.**

**Keywords:** covalent organic framework, high-throughput screening, structure-property, adsorption, cross-attention

## INTRODUCTION

The identification of optimal porous materials for gas adsorption and separation is a central challenge in materials chemistry and chemical engineering: practical applications from greenhouse gas capture to hydrogen purification demand adsorbents that combine high capacity, strong selectivity, facile regenerability and adequate kinetics. Crystalline porous materials are distinguished by their high crystallinity, permanent porosity, diverse pore architectures, tunable pore sizes, and adjustable chemical composition; these combined features provide the structural and chemical versatility needed for applications such as gas storage [1,2], molecular separation [3–7], catalysis [8,9], and sensing [10–14]. COFs are a particularly attractive class because modular synthesis permits systematic tuning of backbone topology, pore geometry and chemical functionality [15–17], which in turn governs adsorption behavior through the interplay of confinement-enhanced van der Waals and capillary forces along with specific host-guest interactions mediated by pore-wall functional groups (*e.g.*, hydrogen bonding, dipole and electrostatic interactions) that jointly determine capacity and selectivity [18,19]. Yet the COFs design space is enormous combinatorial choices of building blocks, linkages and nets generate far more candidates than can be assessed experimentally or by brute-force simulation motivating large curated and hypothetical databases [20–22] and high-throughput computational screening (HTCS) efforts [23]. Because rigorous Grand Canonical Monte Carlo (GCMC)-based HTCS remains costly at very large scale, surrogate and machine-learning (ML)-assisted workflows have emerged to accelerate discovery by trading some generality for throughput. This trend is not unique to COFs but pervades the broader crystalline-materials community, motivating ML-assisted high-throughput screening across diverse crystal classes. Combining HTCS with machine learning therefore offers a practical route to screen expansive COFs spaces efficiently and to prioritize candidates for higher-fidelity simulation or experiment [24–29].

It is well-established that structure fundamentally determines functionality; the Crystallographic Information File (CIF) of COFs inherently contains all information regarding their properties. However, predicting structure-property relationships for crystalline materials such as COFs has consistently proven to be a formidable challenge, since models struggle to learn a reliable mapping from inputs to these derived outputs. Archived research often incorporates gas-specific descriptors computed from molecular simulations such as Henry coefficients or adsorption heat, either as model features or as prescreening criteria. This approach, however, poses two critical risks: first, these descriptors implicitly encode particular gases and thermodynamic conditions (including pressure, temperature, and force-field assumptions), limiting the models' transferability to other adsorbates or operating regimes; second, computing these descriptors is computationally expensive, undermining scalability. Concrete studies illustrate this limitation: Gokhan Onder Aksu *et al.* integrated GCMC simulations with ML to predict COFs gas adsorption/separation performance, but their models consistently relied on GCMC-derived gas-specific features (*e.g.*, adsorption heat for $CH_4/H_2$ separation [30], Henry coefficients for $CO_2/CH_4$ separation and single-component uptake [31,32]); similarly, De Vos *et al.* (GCMC-ML screening [28]) and Qiu *et al.* (CDFT-string method-ML framework for $CH_4/H_2$ separation [33]) also depended on gas-specific parameters (*e.g.*, Henry coefficients) from simulations, resulting in high computational costs. Notably, to break free from this reliance on gas-specific descriptors, some studies have attempted to use other data processing methods. However, such attempts have suffered from poor predictive performance, largely due to inherent flaws in their data handling: these methods often rely solely on structural descriptors calculated by Zeo++ [34], which overlooks crucial geometric and topological features; even when focusing on structural representations, they fail to incorporate chemical principles. Both issues prevent the capture of multifaceted structure-property relationships, a key factor for accurate prediction [35]. Multi-modal fusion models have been widely applied in materials research. Existing studies have demonstrated that the use of multi-modal feature fusion can combine local and global features of data, thereby expanding the usable feature space of the model and improving its generalization ability [27,40]. For broad, deployment-relevant screening, it is therefore preferable to learn compact, transferable structure-property mappings that are driven primarily by the frameworks' geometry and chemistry.

[1] College of Science, China Agricultural University, Tsinghua East Road 17, Beijing 100083, China
[2] College of Information and Electrical Engineering, China Agricultural University, Tsinghua East Road 17, Beijing 100083, China
[3] College of Environmental Science and Engineering, North China Electric Power University, Beinong Road 2, Beijing 102206, China
[4] Qingdao Institute of Software, College of Computer Science and Technology, China University of Petroleum (East China), West Changjiang Road, Qingdao, Shandong 266580, China
[5] College of Chemistry and Chemical Engineering, Weifang University, Weifang, Shandong 261061, China
[*] Corresponding author (email: zschen@ncepu.edu.cn(Chen Z); xkwang@ncepu.edu.cn(Wang X); feifanzhang@cau.edu.cn(Zhang F))

Previous research limitations stemmed from incomplete extraction of complex structural information embedded in pristine crystal frameworks. To address this challenge, we have systematically explored diverse mathematical methodologies, integrated cutting-edge concepts from protein-related research, leveraged artificial intelligence, and then from an interdisciplinary perspective, we propose a novel methodological framework designed for COFs Adsorption Prediction (COFAP). To comprehensively and accurately capture the structural characteristics of COFs and overcome the limitations of single-modal representation, this study introduces multiple sets of complementary modalities. The cross-sectional image modality reduces 3D COFs to interpretable 2D representations, balancing computational efficiency and the integrity of structural information; the dual modalities of persistent homology and Zeo++ provide topological fingerprints and macrostructural parameters, respectively, improving the model's perception ability and prediction accuracy; the coarse-grained bipartite supragraph modality eliminates atomic redundancy, focuses on the core of linker-linkage chemistry, and provides support for the model to learn structure-property relationships. Workflow of the whole research (shown in **Fig. 1** (A)) comprises four main stages: (1) Data acquisition. The study uses the hypoCOFs [36] collection of 69,840 computationally generated COFs structures, with property labels ($CH_4$ uptake at 0.1, 1 and 10 bar; $H_2$, $CO_2$, $N_2$, $O_2$ uptake at 1 bar) generated from GCMC simulations [30,32]. Note that we are trying to avoid using gas-specific-related features. (2) Multi-modal feature extraction. As the structural information hidden in CIF is inherently complex and rich, to achieve a comprehensive understanding of COFs, it is essential to extract information from multi perspectives. Three routs of deep learning methods are specifically designed to extract multi-modal features, including basic structural and chemical features (**Fig. 1** (B)), hidden topo structural features (**Fig. 1** (C)), and hidden group chemical features (**Fig. 1** (D)). (3) Cross-modal feature fusion. The features extracted from a single perspective remain one-sided unless they are integrated. However, arbitrary fusion may lead to adverse effects. Considering that different features have different levels of interpretability and subsequently lead to different priorities among each other, we leverage cross attention mechanism to achieve effective cross-modal information synergy. (4) Screening. Based on COFAP, we obtain the performance ranking of all involved COFs. But in different prediction tasks for adsorption and separation, researchers may focus on different properties. Therefore, we also propose a weight-adjustable sorting method, by which the optimal COFs structures that meet various research goals can be screened out.

## RESULTS AND DISCUSSION

### Multi-modal feature extraction

**Sectional Plane - convolutional Variational AutoEncoder (SP-cVAE).** In gas adsorption and separation, pore geometry plays a decisive role in governing performance, as well as the number and spatial arrangement of different atoms. To capture these key structural and chemical characteristics in a concise and interpretable way, we introduce a sectional plane method that slices COFs supercells along representative crystallographic directions, and projects four atom types (C, H, O, and N) and chemical bonds into two channels within each slice onto 2D planes (**Fig. 2** (a)).

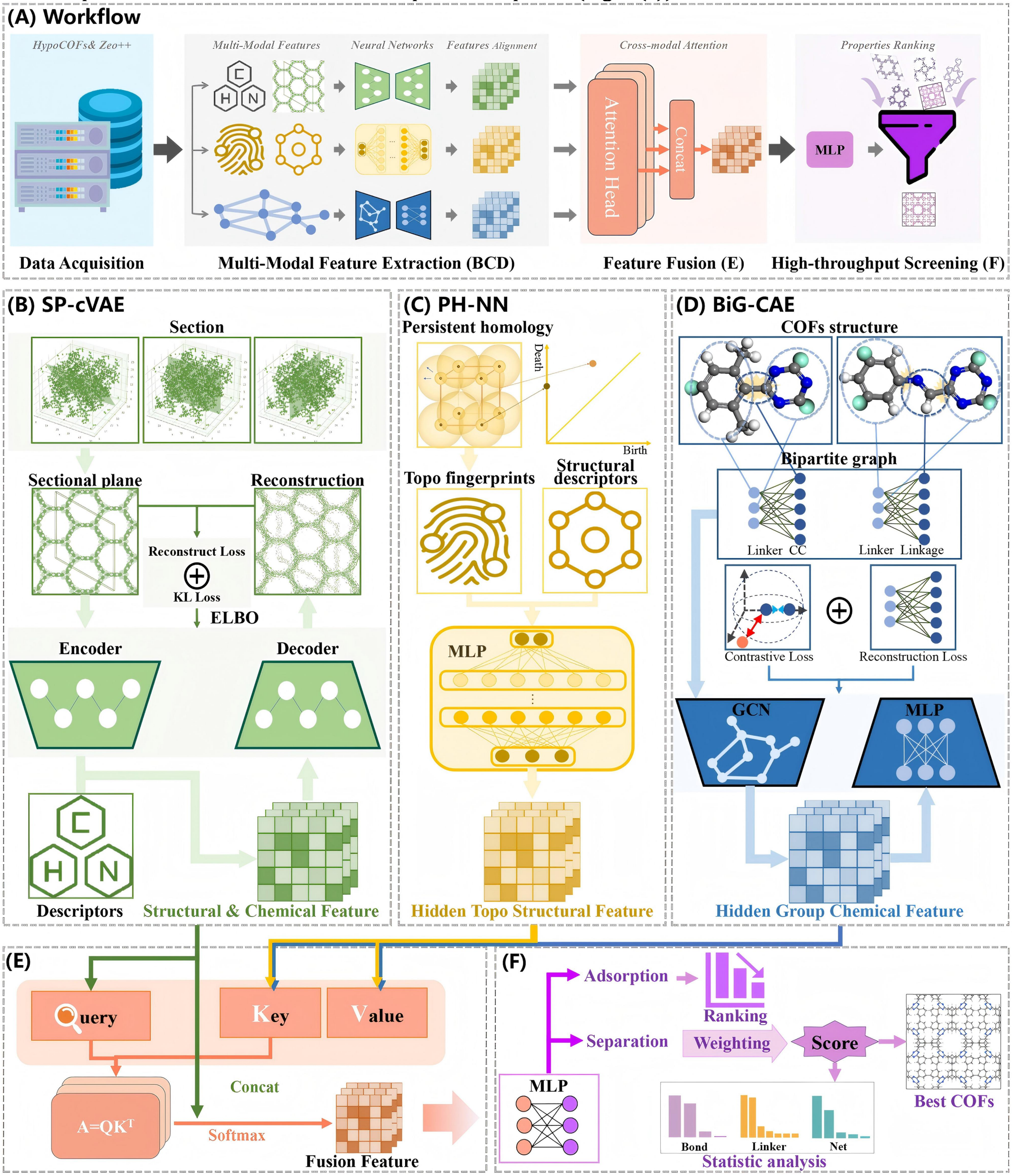

**Figure 1** (A) Overall workflow. (B) SP-cVAE, learns compact structural–chemical representations from sectional planes and global descriptors. (C) PH-NN, extracts topological structural features using persistent-homology fingerprints with global structure descriptors. (D) BiG-CAE, captures group-level chemical representations from coarse-grained bipartite graphs via contrastive and reconstruction learning. (E) Feature fusion, integrates cross-modal representations through attention-based fusion. (F) High-throughput screening, applies COFAP to adsorption and separation tasks to identify high-performing hypoCOFs.

A convolutional variational autoencoder is employed to compress the nine 2D planes into compact latent descriptors that summarize both global pore features and chemical patterns. The model uses a convolutional encoder that outputs the mean and log-variance of a Gaussian distribution for latent-vector sampling, together with a transposed-convolution decoder optimized to reconstruct the atomic-density maps. The nine latent vectors are then aggregated by a 1D convolutional layer to capture inter-directional structural correlations such as pore alignment across planes. This pipeline therefore yields prior-informed, low-dimensional descriptors that directly reflect structural and chemical motifs relevant to adsorption.

**Persistent Homology - Neural Network (PH-NN).** To capture the 3D topology of COFs pore networks information that is complementary to 2D planes and 1D simple geometric measures, the PH-NN encodes two compact structural modalities: a topological fingerprint derived from persistent homology [37–39] (detailed information provided in Methods section) and a set of global geometric descriptors precomputed by Zeo++, including pore limiting diameter (PLD), largest cavity diameter (LCD), accessible surface area ($S_{acc}$), density ($\rho$) and porosity ($\phi$). In COFs adsorption prediction, the core advantage of persistent homology lies in its ability to accurately capture the intrinsic topological structure and connectivity of pore networks, making up for the shortcomings of traditional geometric descriptors. By constructing Vietoris-Rips complexes and extracting $H_0$ connectivity as well as $H_1$ loop/tunnel features, it converts the pore space of COFs into quantifiable topological fingerprints, which effectively encode key information such as pore connectivity pathways, nested relationships, and spatial distributions—all of which directly determine the diffusion paths of gas molecules and the accessibility of adsorption sites. Its physicochemical significance mainly includes two aspects: first, by recording the "birth-death" process of topological features, it screens out the stable pore structures that dominate adsorption behavior, avoiding the one-sided description of complex pore systems by geometric parameters at a single scale; second, it complements macroscopic structural parameters such as pore diameter and specific surface area, revealing the intrinsic correlation between pore topology and adsorption capacity as well as selectivity. This provides a topological theoretical basis for explaining the differences in adsorption performance of COFs with similar geometric structures, thereby improving the prediction accuracy and generalization ability of the COFAP framework. The outputs of the network are concatenated to form the PH-NN structural descriptor, which is then supplied to the cross-modal fusion stage to enrich the SP representations. The pre-trained model acts as a frozen feature extractor in the fusion model.

**Bipartite Graph - Contrastive AutoEncoder (BiG-CAE).** COFs contain many repeating organic motifs, producing a redundant atomic-level description that is unnecessary for adsorption and separation tasks. Because performance depends mainly on pore geometry and the chemistry of connection motifs rather than every atomic detail, a coarse-grained representation is preferable: it reduces dimensionality, improves interpretability, and highlights adsorption-relevant features. Following recent evidence that fine-grained atomic detail is not essential for adsorption task [40], COFs are represented as a bipartite supragraph whose nodes encode linkages (*n*, *e.g.* imine, amide, CC) and linkers (*l*, the organic building blocks) (**Fig. 2** (b)), and all plausible linker-linkage pairings are initially included (a complete bipartite assembly) to avoid arbitrary assumptions about connectivity, leaving the encoder to learn which connections matter [41]. Importantly, the node features are explicitly chemical, so the encoder extracts hidden group chemical features that complement prior chemical features extracted by SP-cVAE.

After pre-training, the encoder is used as a frozen feature extractor: its latent and hidden representations are incorporated as auxiliary hidden group chemical features, enriching the sectional plane branch in the fusion model.

## Cross-modal feature fusion

To integrate complementary modalities while preserving the integrity of the primary predictor, cross-attention is adopted as the fusion mechanism [42–44]. The SP-cVAE was selected as the primary model for two reasons. First, it yields a comprehensive yet compact representation by jointly extracting structural and chemical signatures; the two auxiliary encoders (PH-NN and BiG-CAE) were specifically chosen to supplement the SP-cVAE with richer structural topology and detailed chemical fragment information, respectively. Second, the SP-cVAE learns low-dimensional, task-relevant features that reduce learning complexity and limit the influence of redundant signals. Whereas, the auxiliary encoders extract higher-dimensional, locally concentrated and more latent feature sets that increase optimization difficulty and risk introducing noisy or spurious correlations. Selecting SP-cVAE as the principal predictor therefore

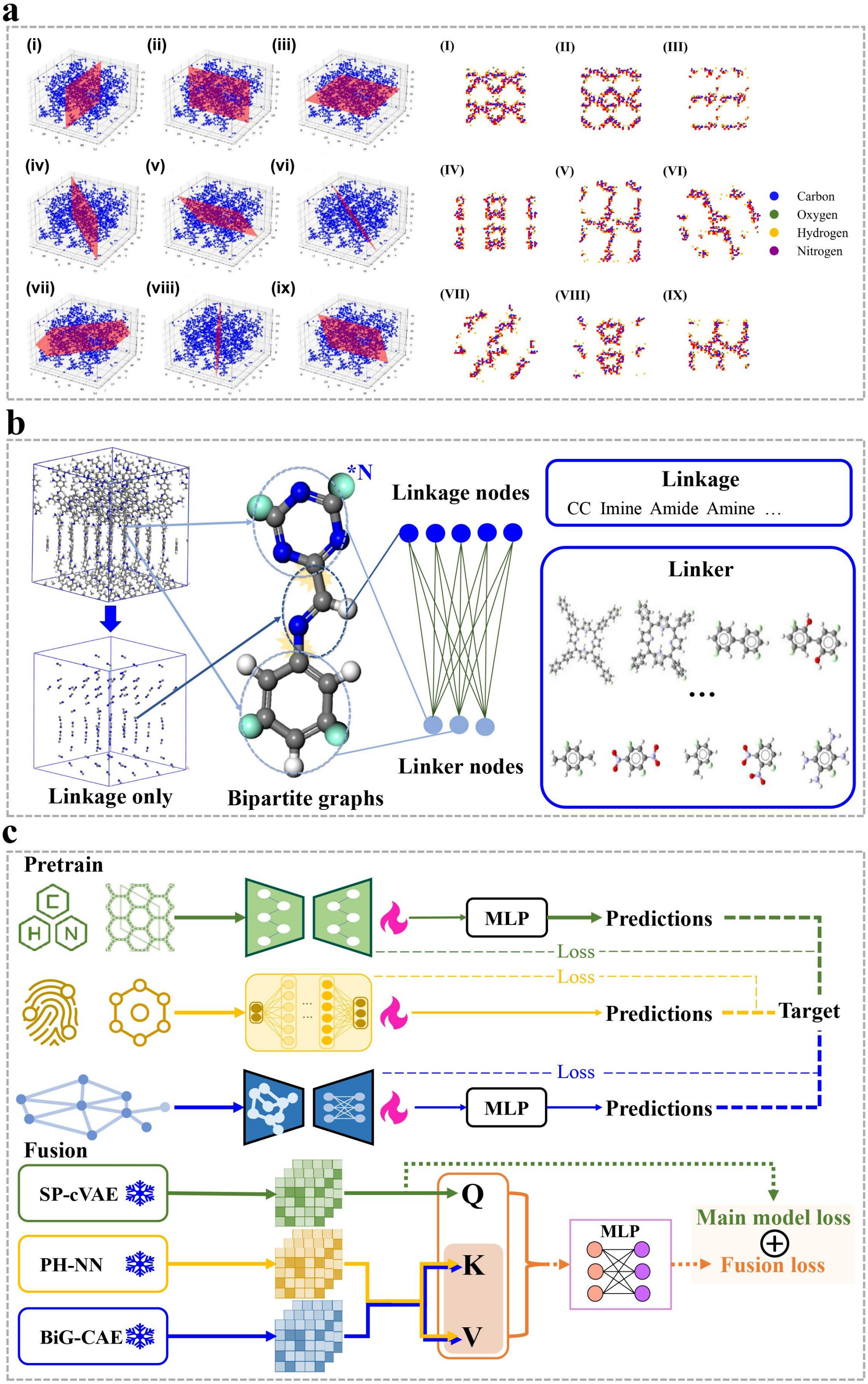

**Figure 2** (a) Illustration of the nine sectional planes that project 3D COF structures into 2D atom–bond views, covering axial, diagonal, and off-axis orientations; example projections (I–IX) correspond to the 3D planes (i–ix). Example shown: linker100_$CH_2$_linker12_NH_qtz_relaxed_interp_2. (b) Construction of bipartite graphs with linkage and linker nodes, where linkage positions are identified via distance-based screening of CIF geometries. (c) Fusion model setup in which pre-trained SP-cVAE, PH-NN, and BiG-CAE encoders are frozen and serve as feature extractors, with SP-cVAE providing queries and the auxiliary branches supplying keys and values for cross-attention. The learning module is formulated as a contrastive autoencoder operating on the heterograph supragraph. The encoder is a heterogeneous graph-convolutional network that hierarchically aggregates node information and pools hidden states into a compact latent vector. The contrastive loss is derived from temperature-scaled cosine similarity, which aligns augmented views of the same COFs and separates distinct COFs in the latent space.

enhances the effectiveness and stability of the fusion stage by ensuring that downstream attention focuses on corroborative auxiliary information, rather than on abundant but less directly informative features.

In this scheme, the SP-cVAE supplies the query ($Q$) while the auxiliary branches supply keys ($K$) and values ($V$) (**Fig. 2** (c)), realizing the scaled dot-product attention computation. All pre-trained weights are frozen to protect learned representations.

## Performance in prediction

The prediction targets include single-component gas uptake ($CO_2$, $H_2$, $N_2$, $O_2$ at 1 bar, 298 k), $CH_4/H_2$ separation performance (adsorption selectivity $S_{CH_4/H_2}$, working capacity $\Delta N_{CH_4}$), and $CH_4$ uptakes under different pressures (0.1, 1, 10 bar). We trained COFAP on these targets. COFAP has the ability to generalize from separation targets to various kinds of gas uptakes and remain stable under pressure variations focused dataset, which highlights its practical value for diverse industrial scenarios and its role as a universal predictive tool in COF-based gas adsorption and separation studies.

**Table 1.** Performance metrics of the COFAP model across three categories of prediction targets.

| Target | $R^2$ | RMSE | MAE | $R$ | $r_s$ |
|---|---|---|---|---|---|
| $S_{CH_4/H_2}$-VSA | 0.9446 (0.0046) | 0.0489 (0.0020) | 0.0341 (0.0008) | 0.9748 (0.0022) | 0.9739 (0.0022) |
| $S_{CH_4/H_2}$-PSA | 0.9226 (0.0326) | 1.7897 (0.4683) | 0.8377 (0.0327) | 0.9634 (0.0159) | 0.9779 (0.0024) |
| $\Delta N_{CH_4}$-VSA | 0.8920 (0.0112) | 0.0632 (0.0019) | 0.0373 (0.0014) | 0.9487 (0.0051) | 0.9478 (0.0031) |
| $\Delta N_{CH_4}$-PSA | 0.8892 (0.0169) | 0.0639 (0.0031) | 0.0378 (0.0010) | 0.9472 (0.0082) | 0.9474 (0.0037) |
| $N_{CH_4}$-1bar | 0.9043 (0.0169) | 0.0686 (0.0069) | 0.0403 (0.0017) | 0.9538 (0.0075) | 0.9505 (0.0070) |
| $N_{H_2}$-1bar | 0.9601 (0.0042) | 0.0018 (0.0001) | 0.0013 (0.0001) | 0.9932 (0.0004) | 0.9944 (0.0003) |
| $N_{CO_2}$-1bar | 0.8346 (0.0258) | 0.3805 (0.0236) | 0.2340 (0.0086) | 0.9167 (0.0166) | 0.8930 (0.0108) |
| $N_{N_2}$-1bar | 0.7940 (0.0070) | 0.4329 (0.0066) | 0.2779 (0.0031) | 0.8944 (0.0049) | 0.8868 (0.0070) |
| $N_{O_2}$-1bar | 0.7941 (0.0181) | 0.4318 (0.0222) | 0.2852 (0.0098) | 0.8935 (0.0097) | 0.8839 (0.0105) |
| $N_{CH_4}$-10 bar | 0.9305 (0.0076) | 0.2636 (0.0159) | 0.1843 (0.0053) | 0.9692 (0.0025) | 0.9673 (0.0023) |
| $N_{CH_4}$-0.1 bar | 0.8742 (0.0231) | 0.0112 (0.0012) | 0.0058 (0.0003) | 0.9398 (0.0099) | 0.9313 (0.0076) |

Beyond value accuracy metrics ($R^2$, MAE and RMSE), we evaluated COFAP for its ability to reproduce material rankings and inference efficiency. As ranking consistency between model predictions and ground truth is critical for screening, it was quantified using Pearson and Spearman correlation coefficients (definitions and formulas of the above 5 statistical metrics can be found in Table S1). To assess practical applicability for large-scale COFs screening, we measured inference throughput on an NVIDIA GeForce RTX 4090 using the hypoCOFs library (69,840 structures). COFAP performs excellently and consistently on seen and unseen data, demonstrating strong generalization: $R^2$, Pearson and Spearman correlation coefficients for most metrics exceed 0.9, indicating the model captures not only absolute values but also relative material rankings important for screening. Moreover, the measured inference speed averaged $158 \pm 30$ samples $s^{-1}$, a throughput that far outpaces methods requiring per-structure Widom insertion or GCMC calculations (*e.g.*, adsorption heat or Henry coefficients), and thus offers a clear advantage for high-throughput discovery workflows, the complete metrics is shown in Table 1. The scatter plot group for the $CH_4/H_2$ separation task is shown in **Fig. 3** (a), and the remaining scatter plots are presented in Figs. S1-S3.

**Table 2.** Comprehensive comparison of model performance across different prediction tasks.

| Metrics | Model | $S_{CH_4/H_2}$ VSA† | $S_{CH_4/H_2}$ PSA† | $N_{CH_4}$ 10 bar | $N_{CH_4}$ 1 bar | $N_{CH_4}$ 0.1 bar | $N_{CO_2}$ 1 bar |
|---|---|---|---|---|---|---|---|
| $R^2$ | This study | **0.9402** | **0.9028** | **0.9294** | **0.9066** | **0.8252** | **0.8756** |
| | Reference[30] | 0.8680 | 0.8830 | 0.6270 | 0.6170 | 0.4640 | – |
| | Reference[31] | – | – | 0.9280 | 0.6880 | – | 0.6130 |
| | Kernel Ridge | 0.7652 | 0.8055 | 0.7486 | 0.6454 | 0.5048 | 0.7969 |
| | Random Forest | 0.7621 | 0.8205 | 0.7517 | 0.6252 | 0.5032 | 0.8583 |
| | XGBoost | 0.7671 | 0.8227 | 0.7867 | 0.6638 | 0.4918 | 0.8620 |
| RMSE | This study | **0.0484** | **1.7824** | 0.2538 | **0.0111** | 0.1872 | 0.3056 |
| | Reference[30] | 3.3500 | 2.6100 | 0.6200 | 0.1500 | 0.0300 | – |
| | Reference[31] | – | – | **0.1330** | 0.0540 | – | **0.2350** |
| | Kernel Ridge | 4.4580 | 3.1908 | 0.5103 | 0.1430 | **0.0253** | 0.4477 |
| | Random Forest | 4.4871 | 3.0650 | 0.5071 | 0.1470 | **0.0253** | 0.3739 |
| | XGBoost | 4.4397 | 3.0462 | 0.4701 | 0.1393 | 0.0256 | 0.3691 |
| MAE | This study | **0.0355** | 1.0813 | **0.0111** | **0.0066** | **0.0066** | **0.0422** |
| | Reference[30] | 1.2800 | **1.0600** | 0.4800 | 0.1000 | 0.0100 | – |
| | Reference[31] | – | – | 0.1330 | 0.0300 | – | 0.1060 |
| | Kernel Ridge | 1.9895 | 1.2988 | 0.3585 | 0.0834 | 0.0112 | 0.2965 |
| | Random Forest | 1.8768 | 1.2149 | 0.3559 | 0.0809 | 0.0110 | 0.2615 |
| | XGBoost | 1.8382 | 1.1658 | 0.3396 | 0.0796 | 0.0112 | 0.2608 |

**Bold**: overall best.

†: Since reference [30] did not provide prediction metrics for the model without adsorption heat features, the model with adsorption heat is used here for comparison.

**Ablation study.** To verify the necessity and contribution of each modal in COFAP, we performed ablation studies, including SP-cVAE, PH-NN and BiG-CAE, which is separated into CC and non-CC, as the node ($n$) of the structures whose linkers are directly connected by carbon atoms differs from those connected by linkages, and the fused COFAP model itself (configuration provided in Table S2). The experimental protocol for all components remained consistent: each modal was trained independently on the same unseen COFs dataset and evaluated on the same set of prediction tasks. Performance was compared using the same metrics ($R^2$, RMSE, MAE) to clarify the role of each component in the multi-modal fusion framework. The graphic results of the ablation studies of $R^2$ are shown in **Fig. 3** (b), while the full results are shown in Tables S3-S5.

Among them, SP-cVAE achieved relatively good single-task performance. The PH-NN and BiG-CAE components, though not outstanding in individual training, enabled the fusion model COFAP to outperform any single component in all tasks (achieving higher $R^2$ and lower RMSE and MAE). This indicates that the extracted multi-modal features have good complementary effects, and that the modal fusion performed by COFAP can correctly process the useful information of each modal. Therefore, the robustness and generalization ability of the model is enhanced.

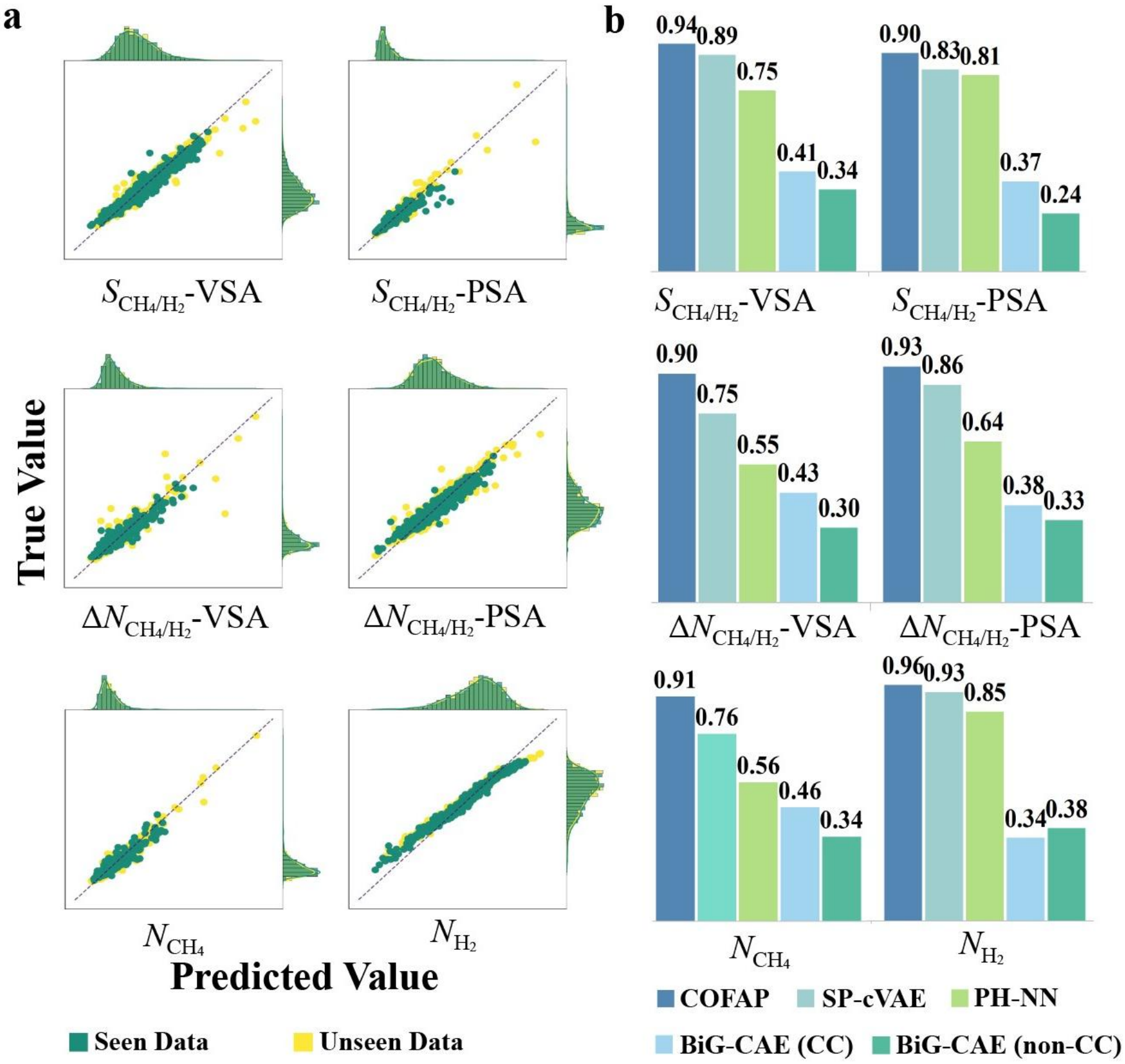

**Figure 3** (a) Scatter plot of unseen data (green) and seen data (yellow) for $CH_4/H_2$ separation task-related targets prediction, where the scatter points are tightly distributed along the diagonal, indicating good predictive performance of the model. (b) The bar charts of ablation study results showing the $R^2$ of model components SP-cVAE, PH-NN, BiG-CAE (which is separated into CC and non-CC, as the node (n) of the structures whose linkers are directly connected by carbon atoms differs from those connected by linkages) and COFAP in predicting the same set of targets as (a). The rest of scatter plots and bar charts are presented in Figs. S1-S3 and Fig. S4 respectively.

And we note an anomalous, large RMSE value for the BiG-CAE-CC on $CH_4$ at 10 bar reported in Table S5; targeted investigation shows that this behavior is not caused by a numerical reporting error, training instability, or feature collapse. Specifically, the corresponding MAE (5.1229) and $R^2$ (0.1335) indicate an overall degraded fit on this particular test subset rather than a single spurious entry. Further analysis reveals that the inflated RMSE is driven by a small number of extreme residuals (outliers), to which RMSE is particularly sensitive, and by differences in label distributions between the CC and non-CC evaluation subsets, which are not identical. For example, the CC test subset for uptake of $CH_4$ at 10 bar exhibits a larger dynamic range (maximum 11.59 mol·kg$^{-1}$, 99$^{th}$ percentile 7.48, standard deviation 1.05, and IQR 1.32) compared with the non-CC subset (maximum 7.82 mol·kg$^{-1}$, 99$^{th}$ percentile 6.27, standard deviation 0.91, and IQR 1.11), explaining why RMSE of BiG-CAE-CC is much larger than that of non-CC, even though they are almost at the same $R^2$ (0.1335 and 0.1311). Importantly, despite the localized performance degradation observed for the BiG-CAE-CC component under specific conditions, the fused COFAP model remains consistently robust across gases and pressures, supporting its improved generalization ability.

**Performance comparison.** The performance of COFAP was evaluated by benchmarking it against established models reported in the literature. Specifically, reference [30] and reference [31] provide machine learning models designed for separation tasks of $CH_4/H_2$ and $CH_4/CO_2$, respectively. The compared targets include: (1) $CH_4/H_2$ selectivity under VSA (Vacuum Swing Adsorption) and PSA (Pressure Swing Adsorption); (2) gas uptakes under typical pressure conditions (10 bar $CH_4$, 1 bar $CH_4$, 0.1 bar $CH_4$, 1 bar $CO_2$). To maintain consistency, we used the same evaluation metrics (Table 2).

For $S_{CH_4/H_2}$, COFAP generally maintained a significant advantage even compared to the model with adsorption heat input. (The model performance without adsorption heat input isn't reported in reference [30]) COFAP performed better on all three targets for $S_{CH_4/H_2}$-VSA. The model from reference [30] only had a slight advantage in MAE for $S_{CH_4/H_2}$-PSA, but this did not diminish the overall superiority of COFAP.

In the gas adsorption task, COFAP's performance remained strong under most pressure conditions. For 10 bar $CH_4$ adsorption, reference [31] achieved results close to COFAP in $R^2$ and RMSE, but COFAP still led in MAE. For 1 and 0.1 bar $CH_4$ adsorption, COFAP outperformed both reference models in all three targets. For 1 bar $CO_2$ adsorption, COFAP outperformed reference [31] in each target. These comparisons confirm that COFAP's performance is significantly better than machine learning models without gas-specific features, and even surpasses models with such features (the model of reference [30]) in adsorption selectivity tasks. This strongly validates that COFAP can discard gas-specific features while maintaining high accuracy, making it very reliable for high-throughput screening applications.

## Application of COFAP on high-throughput screening

COFAP was then deployed in inference mode across the full hypoCOFs collection (69,840 computational structures) to predict single-component gas uptakes at 1 bar for five common adsorbates ($CH_4$, $H_2$, $CO_2$, $N_2$, $O_2$). For each gas, per-structure uptake predictions produce a complete, ranking of the entire dataset. These per-gas rankings serve two immediate screening roles: (i) rapid candidate triage by surfacing the most promising COFs for a given target gas, and (ii) a first-pass filter for separation workflows by identifying materials with complementary adsorption profiles across gas pairs (for example, high $CH_4$ uptake coupled with low $H_2$ uptake). Importantly, although the present study focuses on single-component adsorption, these predictions can be readily combined to estimate ideal separation metrics (*e.g.*, ideal adsorption selectivity) under the assumption of independent adsorption, which is a widely used approximation in high-throughput screening of porous materials. All ranking data and single-component uptake predictions are provided in Supporting Information II, enabling downstream analysis for a variety of gas-pair separation scenarios (**Fig. 4** (a)).

For the separation task, a reproducible prioritization pipeline was developed to convert model outputs into a compact, diversified set of candidate COFs for downstream application. The regenerability $R$% and the adsorbent performance score APS (the formulas are provided in Table S6) derived from selectivity and working capacity become two basic metrics for following analysis. The pipeline implements a small number of transparent steps: metric normalization, an interpretable linear composite score, a systematic weight-sensitivity scan, metric contribution-rate reporting, and aggregation of structural statistics among top-ranked entries. We choose an interpretable linear weighted (ratio) aggregation for the composite score because, after normalizing heterogeneous metrics to a common scale, a linear combination offers a transparent, monotonic and easily tunable representation of stakeholder preferences: weights directly encode relative importance, the computation is trivial for large-scale screening, and the resulting scores are straightforward to decompose into metric contributions for diagnostics. To guard against arbitrary weighting, this linear scheme is used together with weight-sensitivity diagnostics so that trade-offs and robust candidates can be identified without relying solely on a single fixed weighting. And its novelty lies in the combination of flexibility, interpretability and reproducibility.

**Table 3.** Top-10 COFs for VSA $CH_4/H_2$ separation under $w_R = 0.5$, $w_A = 0.5$.

| Name | $S_i(w_R, w_A)$ | $rate_{R,i}$ | $rate_{A,i}$ | bond | net |
|---|---|---|---|---|---|
| linker110_C_linker91_C_tfg_relaxed | 0.6165 | 0.1890 | 0.8109 | CC | tfg |
| linker110_C_linker92_C_tfg_relaxed | 0.6112 | 0.1921 | 0.8078 | CC | tfg |
| linker110_C_linker87_C_mdf_relaxed | 0.6066 | 0.3323 | 0.6676 | CC | mdf |
| linker100_C_linker102_C_cda_relaxed | 0.5625 | 0.5454 | 0.4545 | CC | cda |
| linker102_C_linker100_C_cda_relaxed | 0.5562 | 0.5455 | 0.4544 | CC | cda |
| linker92_C_linker92_C_bpi_relaxed | 0.5489 | 0.4354 | 0.5645 | CC | bpi |
| linker110_C_linker94_C_jeb_relaxed | 0.5337 | 0.9368 | 0.0631 | CC | jeb |
| linker92_C_linker92_C_bpe_relaxed | 0.5318 | 0.5375 | 0.4624 | CC | bpe |
| linker105_C_linker92_C_lil_relaxed | 0.5123 | 0.8837 | 0.1162 | CC | lil |
| linker91_C_linker91_C_qtz-f_relaxed_interp_2 | 0.5076 | 0.2680 | 0.7319 | CC | qtz-f |

This design delivers three practical advances. First, the weight-adjustable composite scoring lets stake-holders tune the ranking to different application priorities (*e.g.* $R$% versus APS) while preserving a stable, reproducible selection procedure. Second, the weight-sensitivity and contribution-rate diagnostics expose when top candidates are robust to weight choices and they reflect strong trade-offs, enabling defensible decision-making instead of opaque ranking. Third, by exporting full, machine-readable ranking matrices and condensed structural summaries, the pipeline supports rapid, diverse candidate nomination for targeted high-fidelity simulation or experiment, and facilitates community reuse. Together these features make the prioritization layer a practical bridge from COFAP predictions to actionable materials discovery.

For instance, industrial practitioners focusing on cyclic operation may assign higher importance to $R$%,

whereas researchers optimizing adsorption capacity and selectivity may emphasize the APS metric. This distinction reflects several practical considerations. In large-scale, continuous or semi-continuous adsorption processes PSA/VSA units, high $R\%$ directly impacts operational expenditure and plant availability: materials with low $R\%$ require more frequent thermal or pressure regeneration, incur higher energy costs, and accelerate bed replacement or refurbishment schedules.

In such contexts, a heavier weight on $R\%$ favors adsorbents that combine adequate uptake with low regeneration penalty, long cycle life, and mechanical/chemical stability under repeated swing conditions. Conversely, laboratory-scale demonstrations, proof-of-concept separations, or single-pass purification tasks often prioritize absolute separation performance and working capacity; here, a higher weight on APS is appropriate because these settings value peak selectivity and per-cycle throughput over long-term cyclic durability (**Fig. 4** (a)).

To illustrate practical implications of the weighting scheme, three representative weight combinations were selected for detailed screening and aggregate reporting under VSA conditions as examples: $w_R : w_A =$ 0.5 : 0.5, 0.2 : 0.8, and 0.8 : 0.2. The first setting corresponds to a neutral (mathematical) average that treats $R\%$ and APS with equal importance; the second emphasizes APS, reflecting laboratory or single-pass high-selectivity use cases; and the third prioritizes $R\%$, reflecting continuous, cyclic industrial operation where energy and cycle life dominate process economics. For each weighting, the pipeline outputs top-10 candidate lists, metric contribution-rates, and aggregate top-100 structural statistics of bond type, net and linker frequencies. Top-10 candidate lists for these three weightings are shown in Tables 3, S7 and S8, and aggregate top-100 structural statistics are presented in **Fig. 4** (b). The best structures under the example conditions are shown in **Fig. 5** (a, b). The Top-10 candidate lists for rest conditions under VSA and PSA are shown in Tables S9-S19 and Tables S20-S30. The rest of aggregate top-100 structural statistics are presented in Figs. S5 and S6. And the best structures for all conditions are shown in Fig. S7. For each weight pair, aggregate statistics are computed over the top-100 candidates to characterize common structural motifs. The following counts are recorded and exported: (i) bond-type frequency, (ii) topology net frequency, and (iii) linker frequency, as shown in **Fig. 4** (c).

To quantitatively assess the sensitivity of candidate rankings to weight variations, we performed a systematic sensitivity analysis across 11 weight combinations (APS-VSA weight ranging from 0 to 1 in 0.1 increments). The overall ranking stability index was calculated as 0.7792 (on a 0-1 scale), indicating good stability of candidate rankings against weight perturbations. The average Jaccard similarity between adjacent weight pairs was $0.7606 \pm 0.1836$, demonstrating substantial overlap ($\sim$ 76%) of top-performing candidates across neighboring weight settings. Furthermore, the Kendall Tau rank correlation coefficient averaged $0.5523 \pm 0.5377$, confirming moderate positive correlation in ranking order across different weight configurations.

Based on consistency scores derived from ranking frequency and variability, we identified the most stable candidates that maintain robust performance across diverse weight settings (Table S31). These high-stability candidates, characterized by low ranking standard deviations and high consistency scores, represent reliable choices for subsequent experimental or simulation validation regardless of specific weight preferences.

We performed a parallel sensitivity analysis for the PSA dataset to ensure comprehensive assessment across different adsorption scenarios (Table S32). The overall ranking stability index for PSA was calculated as 0.8879, indicating even higher stability compared to the VSA analysis. The average Jaccard similarity between adjacent weight pairs was $0.8051 \pm 0.2556$, demonstrating $\sim$ 81% overlap of top candidates across neighboring weight settings. The Kendall Tau rank correlation coefficient averaged $0.7910 \pm 0.2535$, confirming strong positive correlation in ranking order across different weight configurations for the PSA dataset.

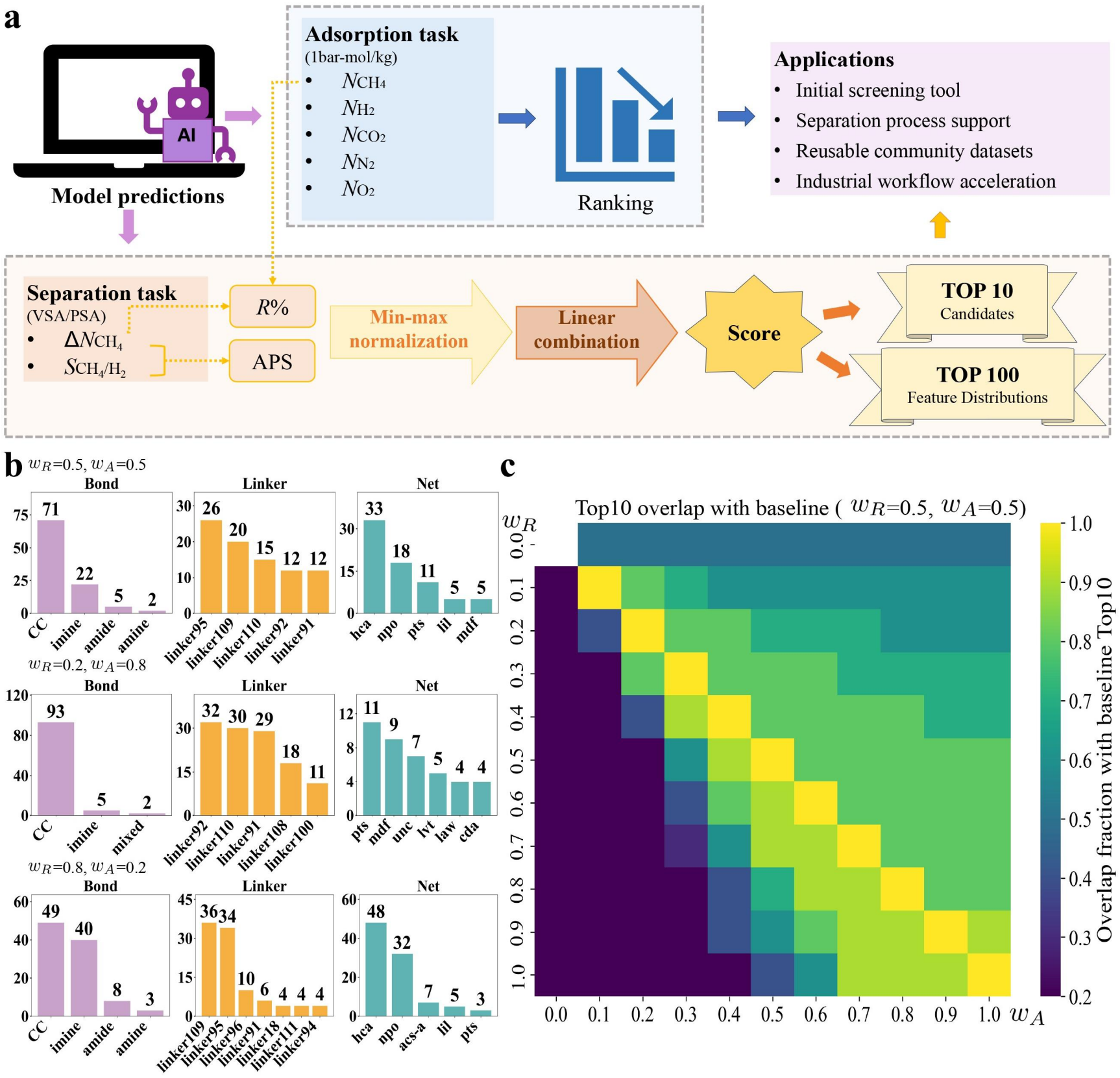

**Figure 4** (a) High-throughput screening workflow, where adsorption rankings and separation-stage composite scores—formed by combining normalized *R%* and APS—yield Top-10 candidates under different weight settings (Tables 3, S7 and S8) and enable aggregation of Top-100 structural statistics. (b) Aggregated linker, linkage, and topology distributions of the Top-100 COFs for $CH_4/H_2$ VSA separation under three weight combinations ($w_R = 0.5$, $w_A = 0.5$; $w_R = 0.2$, $w_A = 0.8$; $w_R = 0.8$, $w_A = 0.2$). (c) Weight-sensitivity heatmap showing Top-10 overlap relative to the baseline ($w_R = w_A = 0.5$), highlighting robust and trade-off-dependent regions. The aggregate top-100 structural statistics of bond type indicate that, with increasing $w_A$, the number of imine rises markedly. The number and spatial distribution of imine groups enhance the material's selectivity toward gas separation, as the lone-pair electrons on imine nitrogen atoms influence the electronic distribution of the framework and thus contribute to selective adsorption of different gas molecules.

Our structural enrichment analysis of the Top-100 candidates ranked by APS under both VSA and PSA conditions reveals a consistent and chemically meaningful pattern: COFs dominated by direct C–C linkages are markedly overrepresented (*e.g.*, 96/100 in the VSA Top-100 versus ~ 30.3% in the full library, fold-enrichment ≈ 3.17; PSA Top-100 shows a similar enrichment), and several specific nets and linkers recur with very high enrichment scores, indicating that particular topology-linker combinations favor superior $CH_4/H_2$ separation. Notably, the Top-100 ensembles are characterized by much smaller pore sizes and cavities (typical PLD ~ 5.2 Å and LCD ~ 7.6 Å in the Top-100 versus ~ 20–25 Å in the full set), lower specific surface area and porosity (Top-100 mean surface area ≈ 3,099 $m^2 \cdot kg^{-1}$ and porosity ≈ 0.198 versus library means ≈ 6,508 $m^2 \cdot kg^{-1}$ and 0.716), and substantially higher framework density (Top-100 mean ≈ 0.825 $g \cdot cm^{-3}$ versus ≈ 0.238 $g \cdot cm^{-3}$ overall). Elemental composition trends further support a chemistry-based origin of the selectivity: Top-100 materials show modestly elevated carbon content and a pronounced enrichment of sulfur-containing motifs (S enrichment fold ≈ 9.8), while particular linker fragments appear repeatedly among high-ranking entries. Taken together, these observations are consistent with a tentative mechanistic picture in which narrow, quasi-microporous cavities that are similar in scale to the kinetic diameter of $CH_4$ could enhance $CH_4$ retention, while carbon-rich (often aromatic) backbones and polarizable heteroatoms (*e.g.*, S) may augment dispersive or inducible interactions; topology and linker chemistry therefore appear to be important structural factors that warrant further targeted investigation.

The inference campaign yields several practical advantages for high-throughput single-component screening and downstream selection. First, surrogate predictions are orders of magnitude faster than structure-by-structure molecular simulation, enabling evaluation of very large libraries in hours rather than months. Second, the full ranked outputs support multi-objective selection without exhaustive simulation like joint consideration of uptake, selectivity and regenerability and integrate naturally with the paper's weight-adjustable prioritization pipeline. Third, the quantitative sensitivity analysis framework provides robust statistical metrics to guide candidate selection, ensuring that chosen materials maintain performance stability across varying optimization preferences. In addition, the complete inference results enable enrichment analysis of top-ranked candidates, revealing chemically interpretable trends in high-performing COFs, including preferences for carbon-rich backbones, specific network/linker motifs, smaller pore apertures, and higher framework densities, which collectively suggest a coupled role of geometric confinement and dispersive host–guest interactions in enhancing $CH_4/H_2$ separation performance. Fourth, predicted adsorption maps facilitate extraction of structure-property trends and the definition of compact pre-screening rules that guide targeted GCMC or experimental validation on a much smaller candidate set. Finally, providing the complete predicted dataset (rankings plus structural descriptors and sensitivity metrics) promotes community reuse and practical adoption in industrial screening pipelines by delivering fast, interpretable metrics for synthesis and process planning.

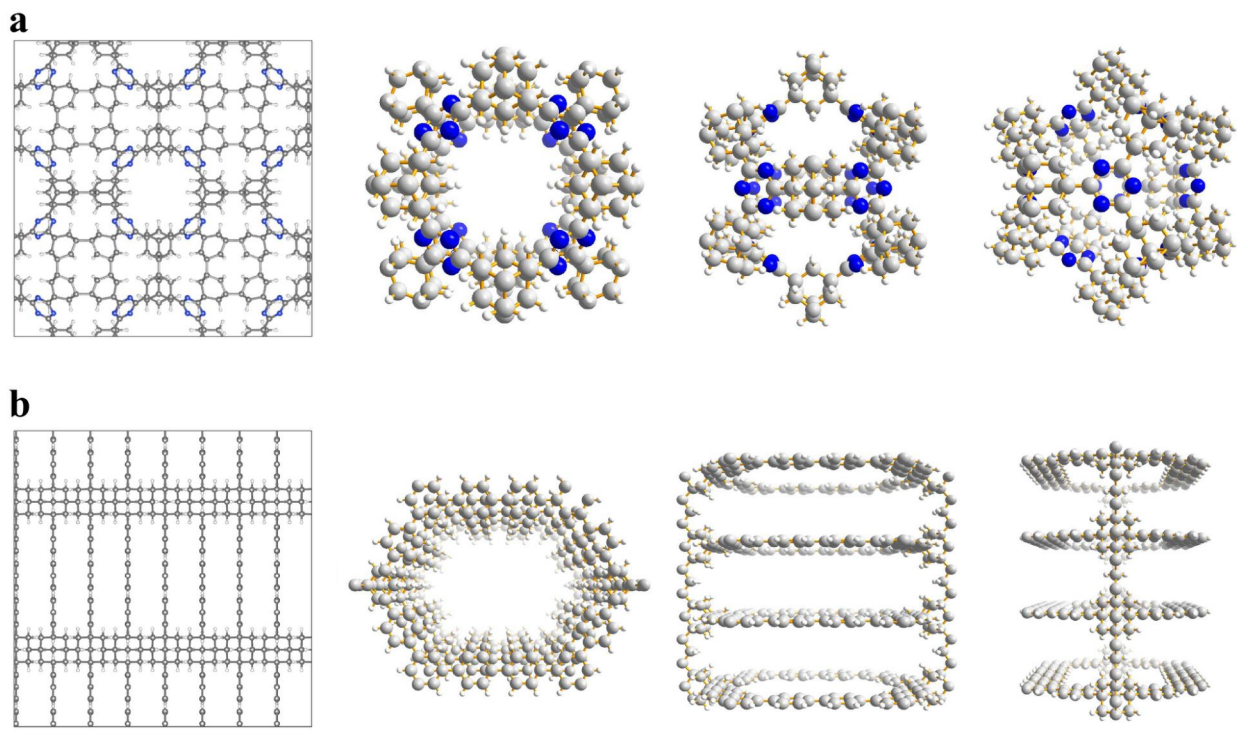

**Figure 5** (a) Visualization of the best COF linker110_C_linker91_C_tfg_relaxed identified in Tables 3 and S7. (b) Visualization of the best COF linker110_C_linker94_C_jeb_relaxed identified in Table S8.

## Stability analysis and improved criteria for pre-screening

The trained framework extracts a heterogeneous set of structural and chemical descriptors namely multi-channel projected sectional planes, persistent-homology topological fingerprints, and coarse-grained linker-linkage connectivity then integrates them through a cross-attention fusion stage to produce final adsorption and separation predictions. Although individual single-modality feature frequently exhibits limited prediction accuracy, the fusion model yields substantially improved and robust performance (as shown in **Fig. 6** (a)). Mechanistically, this improvement arises because the modalities are complementary: persistent homology encodes void connectivity and tunnel structure, sectional planes capture channel alignment and pore-patterns, atomic and elemental descriptors supply local host-guest interaction cues, and the supragraph representation exposes linker linkage motifs that determine the chemical environment of adsorption sites. Cross-attention selectively amplifies corroborating signals across these scales, producing emergent, spatially localized features that correspond to adsorption-active sites descriptors that are difficult to infer from any single input stream alone. This novel multi-modal extraction and fusion mechanism efficiently captures the full hierarchy of crystalline COFs features from pores, channels and spatial physical structure to chemical group distributions, chemistry-related features and adsorption sites. The result is a comprehensive fusion representation that is both chemically interpretable and highly relevant to adsorption/separation behavior, explaining the models' strong empirical performance even when single-modality baselines are weak.

A subsequent statistical analysis of COFAP predictions across the full hypoCOFs dataset identified narrow windows for PLD, LCD, Sacc and porosity $\phi$ (**Fig. 6** (b) for VSA, Fig. S8 for PSA), in which the predicted APS for $CH_4/H_2$ separation is maximized. We adopt APS = 100 mol/kg as the lower-bound threshold for high-performing COFs, for the reason that COFs exceeding this cutoff comprise roughly the top 0.05% of the full datasetclear statistical outliers and the most promising candidates. This threshold yields a moderate-sized, practically manageable subset that reduces computational and synthetic burden while preserving sufficient diversity for downstream computational screening and experimental validation.

The window for PLD is approximately 3.471–6.249 Å and 3.471–6.946 Å for VSA and PSA respectively. This PLD preference can be rationalized from basic adsorption physics. The kinetic diameters of $H_2$ and $CH_4$ are different ($H_2$ ≈ 2.9 Å, $CH_4$ ≈ 3.8 Å) [45]. Therefore, pore windows in the lower end of the identified range are sufficiently large to admit both molecules while remaining tight enough that host-guest van-der-Waals and dispersion interactions selectively favor the larger, more polarizable $CH_4$ ($CH_4$ has a substantially larger static polarizability than $H_2$). As PLD increases, the accessible pore volume and thus working capacity typically grow, which raises APS up to a point. Beyond the upper end of the window, however, pores become so large that specific host-guest interaction strengths weaken (the adsorbate experiences a more bulk-like environment and dispersion contacts are less effective), causing the selectivity component of APS to fall because both gases are accommodated with similar energetics. This mechanistic is consistent with widely reported empirical values and adsorption intuition: typical adsorbate framework contact distances and dispersion-dominated interaction ranges fall in the 3.0–5.0 Å regime (comparable to sums of van-der-Waals radii), and methanes larger polarizability amplifies its dispersion binding relative to hydrogen.

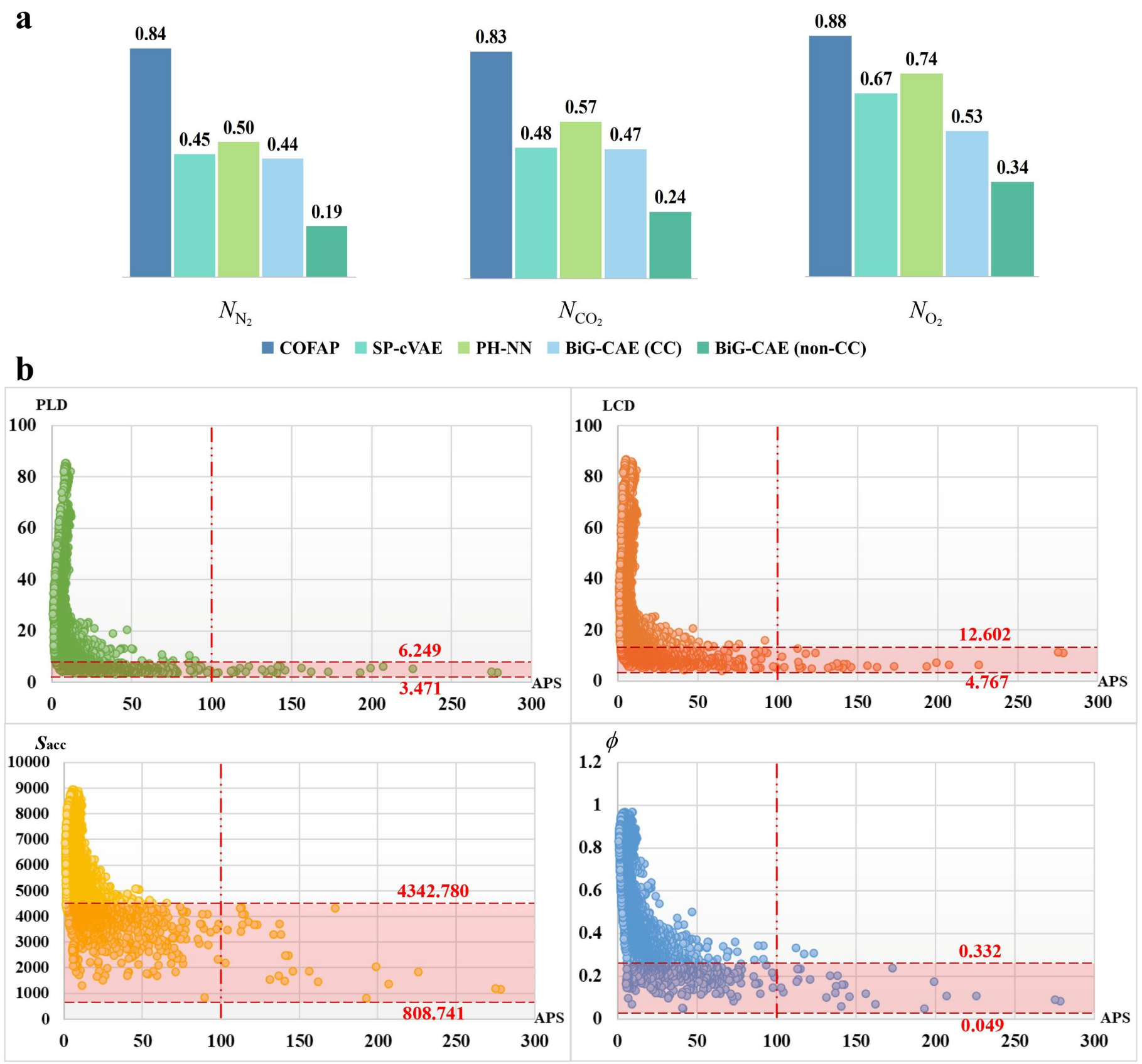

**Figure 6** (a) Bar-chart ablation results for three representative prediction targets, which demonstrate COFAP's strong performance even when single-modality baselines underperform. (b) Statistical scatter plot of PLD, LCD, $S_{acc}$ and porosity $\phi$ versus APS. The plot reveals that high-performing COFs for $CH_4/H_2$ separation under VSA concentrate within a set of narrow window (red-shaded region), highlighting the structural range associated with optimal separation performance.

The statistical analysis of COFAP predictions also reveals a relatively narrow LCD window, approximately 4.767–12.602 Å and 4.767–13.128 Å for VSA and PSA respectively. An optimal range of $S_{acc}$ was identified as well. In general, increasing $S_{acc}$ enhances the adsorption capacity for a single gas species, but an excessively large $S_{acc}$ undermines selective separation among different gases. Conversely, a relatively low $S_{acc}$ can enable precise and efficient $CH_4/H_2$ separation; notably, when the material's $S_{acc}$ approaches the values corresponding to the last two data points of the scatter plot for VSA in **Fig. 6** (b), the APS reaches an astonishing 278.55 mol/kg.

The data also exhibit a clear trend with respect to porosity: materials with higher porosity generally facilitate molecular transport and adsorption desorption kinetics, which favors uptake but can dilute selectivity. For selective separation between different gas species, lower porosity tends to be more favorable because reduced porosity accentuates size- and interaction-based discrimination, thereby promoting selective permeation and capture of a target species and producing an effective selective-rejection toward competing molecules.

Taken together, the statistically inferred windows for PLD, LCD, $S_{acc}$ and porosity $\phi$ furnish a compact, actionable pre-screening rule for downstream simulation or experimental campaigns, and they corroborate the chemical plausibility of COFAP outputs. The model thus identifies regimes that balance selectivity and working capacity for $CH_4/H_2$ separations. From the statistical results we infer that high performance arises from the combined effects of pore size, the spatial distribution of adsorption sites, and surface area, with their synergistic interaction governing the ultimate trade-off between selectivity and capacity.

## CONCLUSIONS

This study presents a universal framework for the structure-property predictions of COFs, COFAP, which shows the best performance in multiple prediction tasks including single-component gas uptake ($CO_2$, $H_2$, $N_2$, $O_2$ at 1 bar, 298 k), $CH_4/H_2$ separation performance ($S_{CH_4/H_2}$, $\Delta N_{CH_4}$), and $CH_4$ uptakes under different pressures (0.1, 1, 10 bar). Compared with traditional experiments, molecular simulations and machine learning models that use gas-related features (*i.e.* adsorption heat and Henry coefficients), COFAP is significantly time-saving. Through COFAP, we can evaluate over ten thousand materials per hour. Although the speed of molecular simulations varies with method and hardware, making direct comparison impractical, COFAP is still orders of magnitude faster. Compared with prediction models that do not use gas-related features, COFAP shows overall leading advantages in $R^2$, RMSE, MAE, Pearson correlation coefficient, and Spearman correlation coefficient. Therefore, COFAP is not only efficient but also accurate in COFs adsorption predictions.

The strong capabilities of COFAP rely on the novel design of multi-modal features extraction, and the cross-modal features fusion framework. Multi-modal features are extracted by three totally different routes based respectively on projected sectional planes, persistent-homology topological fingerprints, and coarse-grained linker-linkage connectivity. The three routes are specially designed for extracting holistic structural and chemical features, hidden topo structural features, and hidden group chemical features. We can see that these features are complementary, and therefore cross-attention fusion stage enables the cross-modal synergy of different features. Besides, two of the routes (SP-cVAE and BiG-CAE) adopt self-supervised architectures (variational and contrastive autoencoders, respectively), which can also contribute to COFAP's robustness and strong generalization.

For the convenience of application, we derive the performance rankings of hypoCOFs and introduce a weight-adjustable sorting method, enabling the screening of optimal COFs that align with diverse research objectives. Statistical analysis of COFAP predictions identifies narrow windows of PLD, LCD, $S_{acc}$ and porosity $\phi$ within which the predicted APS for $CH_4/H_2$ separation is maximized. For VSA, the optimal ranges are approximately PLD 3.471–6.249 Å, LCD 4.767–12.602 Å, $S_{acc}$ 808.741–4342.780 m$^2$/g, and $\phi$ 0.049–0.332. For PSA, the corresponding ranges are PLD 3.471–6.946 Å, LCD 4.767–13.128 Å, $S_{acc}$ 1169.75–4381.19 m$^2$/g, and $\phi$ 0.060–0.362. These range furnishes a practical pre-screening filter that both accelerates downstream GCMC/experimental validation and corroborates the chemical plausibility of COFAP's predictions.

The strong transferability demonstrated by COFAP also makes it a potentially transformative tool for analyzing, predicting, and classifying COFs materials, providing new insights into other areas of COFs materials and laying a solid foundation for next-generation COFs informatics.

Neglecting competitive and co-adsorption effects in rapid, idealized screening can produce systematic underestimates of absolute selectivity in multi-component systems; nevertheless, empirical evidence and sensitivity analyses indicate that relative materials rankings are largely preserved when mixture effects are later introduced. Remaining uncertainties center on (i) the fidelity limits imposed by the classical force fields and rigid-framework assumptions underlying the reference GCMC labels, and (ii) the synthesizability of hypothetical frameworks. The latter concern is partially mitigated by prior validation of the structure-generation protocol against experimentally realized COFs (*e.g.*, COF-300 and TAPBPDA), where computed powder X-ray diffraction patterns were shown to closely match experiment [36]. Taken together, these observations argue for a pragmatic, staged discovery path in which large-scale, low-cost model screening is used to nominate candidates that are then subjected to progressively higher-fidelity simulation and experimental validation as appropriate.

Distinct from conventional single-modality predictors, our framework combines geometric and chemical features to learn richer, site-level representations of adsorption and separation behaviors. This multi-modal integration enables state-of-the-art accuracy, better generalization to unseen COFs, and interpretable, transferable descriptors that directly connect to inverse-design and high-throughput screening workflows. Building on these advances, further incorporation of structural dynamics, interfacial effects, or additional chemical domains could extend the framework beyond COFs to all classes of crystalline porous materials, including MOFs, zeolites, and related frameworks. In this way, our approach provides not only a leading predictive tool, but also a versatile foundation for future materials discovery and design across diverse crystalline systems.

Although the COFAP framework established in this study exhibits excellent performance in adsorption predictions of small gas molecules under ideal thermodynamic conditions, the current model is mainly validated based on existing COF structures and a limited set of small-molecule systems, without systematically covering realistic experimental scenarios such as macromolecular systems, more complex gas mixtures, and non-equilibrium states. Therefore, for experimentally synthesized COFs, the predictive reliability of COFAP is affected by dataset quality, force field accuracy, gas component types, and thermodynamic states. In practically synthesized COF materials, this method is more suitable for adsorption and separation systems with well-defined pore structures and typical small-molecule target gases; in contrast, the prediction accuracy may exhibit certain deviations for systems that differ significantly from known COFs in morphology, or for systems under specific thermodynamic states or complex multi-component gas mixtures.

## METHODS

### Data acquisition

This study utilized the hypoCOFs dataset containing 69,840 computationally generated COFs structures, each accompanied by CIF with atomic coordinates and lattice parameters. Structural descriptors (*e.g.*,

PLD, LCD, $S_{acc}$, $\rho$, $\phi$) were extracted using Zeo++.

Our initial research focused on $CH_4/H_2$ separation under PSA and VSA. Adsorption data were derived from GCMC simulations reported by prior research [30,32], conducted via RASPA [46] with the DREIDING force field for frameworks, TraPPE for $CH_4$, and the Buch potential for $H_2$; Lennard-Jones 12-6 potentials and Lorentz-Berthelot mixing rules governed dispersion interactions. Each simulation comprised equilibration followed by production cycles. Although the original study also reported adsorption heat at infinite dilution via the Widom insertion method, only the mixture uptake data were used here. The reason is that there can be target leakage from thermodynamic features that are intrinsically coupled to uptake values, and may prevent the learning of independent structure-property relationships.

Key separation targets, namely $S_{CH_4/H_2}$ and $\Delta N_{CH_4}$, were computed from uptake. To improve data quality by avoiding near-zero dominated distributions and reduce computational overhead, prior research identified optimal structural regimes (LCD $< 20$ Å, $\phi < 0.80$ Å) based on top-performing experiment-based CoRE COF [47], while restricting the search space to 7,743 structures [30]. This pre-screened subset was then used for model development and was divided into a training/validation set of 6,000 COFs (seen) and an independent test set of 1,137 COFs (unseen).

The predictive framework was subsequently extended to additional industrially relevant gases, namely $CH_4$, $H_2$, $CO_2$, $N_2$, and $O_2$, utilizing GCMC-derived uptake data at 1 bar on the same COFs from prior research [32]. In this case as well, the Henry coefficients reported in the source study were not employed, for the same reason as non-using of adsorption heat.

The unseen evaluation set mixes computational hypoCOFs with experimentally characterized CoRE COFs to test stability across simulated and experimental references.

## Multi-modal feature extraction

**SP-cVAE.** The Sectional Plane (SP) method reduced 3D COFs structures to interpretable 2D representations: these directions are selected to cover a wide range of spatial orientations, thereby ensuring comprehensive structural coverage. COFs supercells were sliced into thin slabs along 9 crystallographically diverse directions defined by distinct normal vectors, atoms and bonds within each slab were orthogonally projected onto a 2D plane. This section reduces dimensionality and effectively preserves planar-level structural information, such as the alignment of pore channels, the tiling pattern of aromatic rings, the connectivity between linkers and nodes, and the structural shaping of diffusion pathways by the framework.

For sectional projection generation, each COF supercell was first expanded to ensure sufficient periodic boundary coverage. A slab thickness of 2 Å was employed to capture atomic layers while maintaining projection clarity. Atoms within each slab were orthogonally projected onto the plane perpendicular to the normal vector, with their positions mapped to a 64 × 64 pixel grid spanning the supercell dimensions. The atom channel encoded atomic identities through distinct grayscale values (C: 0.3, N: 0.5, O: 0.7, H: 0.1, other elements: 0.9), preserving chemical heterogeneity. The bond channel represented covalent connections as line segments between projected atomic positions, with bond intensity scaled by bond order (single: 0.5, double: 0.8, triple: 1.0, aromatic: 0.7). This dual-channel representation explicitly encodes both compositional and topological information critical for adsorption property prediction.

Each sectional plane was converted into a fixed-size two-channel image, consisting of an atom channel and a bond channel, where atom types are distinguished by values, forming input tensors for the SP-cVAE. The model architecture included a convolutional encoder $q_\phi(z \mid x)$ and a transposed convolutional decoder $p_\theta(x \mid z)$. The convolutional encoder processes images via 2D convolutional layers to output mean $\boldsymbol{\mu}$ and log variance $\log\boldsymbol{\sigma}$ of a latent Gaussian distribution; latent vectors $z$ are sampled via the reparameterization trick $(\boldsymbol{z} = \boldsymbol{\mu} + \boldsymbol{\sigma} \odot \boldsymbol{\varepsilon}, \boldsymbol{\varepsilon} \sim N(0, \boldsymbol{I}))$, and the transposed convolutional decoder reconstructs atomic density maps from latent vectors.

Training optimized the evidence lower bound (ELBO) to balance reconstruction and regularization [48]:

$$L(\theta, \phi; \boldsymbol{x}) = E_{q_\phi(z|\boldsymbol{x})}[\log p_\theta(\boldsymbol{x} \mid \boldsymbol{z})] - D_{\mathrm{KL}}(q_\phi(\boldsymbol{z} \mid \boldsymbol{x}) \,\|\, p(\boldsymbol{z})) \tag{1}$$

where $E_{q_\phi(z|\boldsymbol{x})}[\log p_\theta(\boldsymbol{x} \mid \boldsymbol{z})]$ is the reconstruction loss (preserves adsorption-relevant structural features) and $D_{\mathrm{KL}}(_\|_)$ is the KL (Kullback-Leibler) divergence which regularizes latent space to follow a standard normal prior $p(\boldsymbol{z})$ by measuring the difference between the encoder's output distribution and the prior, encouraging a smooth, continuous latent space that prevents overfitting and supports meaningful interpolations (hyperparameters in Table S33).

The hyperparameter selection followed a principled approach: the latent dimension of 128 was chosen to balance representational capacity with computational efficiency, capturing essential structural motifs without overfitting. The VAE weight ($\alpha = 0.01$) and KL divergence weight ($\beta = 0.1$) were determined through grid search to optimize the trade-off between reconstruction fidelity and latent space regularization. The 64 × 64 image resolution was selected to preserve atomic-level details while maintaining computational tractability, with each pixel representing approximately 0.1–0.3 Å depending on supercell size. The slab thickness of 2 Å ensures capture of complete atomic layers while minimizing overlap between adjacent sections.

For each COFs, 9 latent vectors, one per section, each 64-dimensional, were aggregated via a 1D convolutional fusion layer to capture inter-directional correlations, like how pore alignments in different planes collectively influence gas transport), then concatenated with the latent vector mean ( $\bar{z}$ ) and a set of scalar chemical descriptors processed by a separate 2-layer MLP. Total loss combined ELBO and regression loss (MAE):

$$L_{\text{total}} = \alpha \cdot L_{\text{ELBO}} + \beta \cdot L_{\text{regression}} \tag{2}$$

where weights $\alpha$ and $\beta$ were chosen to balance the competing objectives of accurate reconstruction and precise prediction. The fusion layer employs two 1D convolutional stages (9→64→32 channels) with 3×1 kernels to capture local correlations between adjacent sectional planes, followed by adaptive average pooling to produce a compact 32-dimensional fused representation that encodes global structural anisotropy.

**PH-NN.** This model captures three-dimensional topological and geometric information using two complementary modalities. The first modality encodes topological fingerprints derived from atomic coordinates through persistent homology, a computational-topology framework that records the appearance and disappearance of topological features as simplices are added to form a filtration [37]. Concretely, we construct a VietorisRips complex with a maximum edge length of 10.0 Å, to extract $H_0$ connectivity and $H_1$ loop or tunnel features. Persistence diagrams are filtered by a minimum-persistence threshold and then vectorized by histogramming birth-death pairs over the [0,5] Å, producing an 18-dimensional topological fingerprint. The second modality comprises global structural descriptors precomputed with Zeo++, including PLD, LCD, $S_{\text{acc}}$, $\rho$ and $\phi$.

Each modality is processed through a dedicated MLP with batch normalization and dropout, and the concatenated hidden representations form the PH-NN descriptor. (Hyperparameters in Table S34).

**BiG-CAE.** COFs were represented as coarse-grained bipartite supragraphs to capture linker-linkage chemistry without atomic redundancy, where nodes are linkage motifs (*n*, *e.g.*, imine, CC) and organic linkers (*l*). Linkage identification is implemented via informative distance-based screening of CIF geometries to locate covalent connection sites, excluding aromatic rings via a dual-criterion procedure combining local neighbor counting and pairwise distance analysis implemented with spatial indexing for computational efficiency. After exclusion of aromatic rings, candidate linkage sites are located by evaluating elemental identity and interatomic distances consistent with known bond motifs.

The model was a contrastive autoencoder with three loss terms:

$$L_{\text{total}} = \beta \cdot L_{\text{contrastive}} + \alpha \cdot L_{\text{reconstruction}} + L_{\text{regression}} \tag{3}$$

where $L_{\text{contrastive}}$ is temperature-scaled cosine similarity to align augmented views of the same structure and separate distinct structures in the projection space:

$$L_{\text{contrastive}} = -\sum_i \log \frac{\exp\left(z_i^{\top} z_i^{+} / \tau\right)}{\sum_j \exp\left(z_i^{\top} z_j / \tau\right)} \tag{4}$$

where $z_i^{\top} z_i^{+}$ are representations of augmented views, and $\tau$ is the temperature parameter; $L_{\text{reconstruction}}$ is Huber loss for faithful decoding of latent representations, defined as:

$$L_{\text{reconstruction}} = L_{\text{Huber}(y,\hat{y})} = \begin{cases} \frac{1}{2}(y-\hat{y})^2, |y-\hat{y}| \le \delta \\ \delta(|y-\hat{y}| - \frac{1}{2}\delta), |y-\hat{y}| > \delta \end{cases} \tag{5}$$

where $\delta$ denoting the transition threshold; $L_{\text{regression}}$ is supervised loss for property prediction (hyperparameters are listed in Table S35).

The encoder is a heterogeneous graph-convolutional neural network that hierarchically aggregates node information and pools hidden states into a compact latent vector via a nonlinear projection. After pretraining, the weights of the encoder are frozen, and then used as a feature extractor during the fusion phase.

## Cross-modal feature fusion

Cross-attention enables selective, data-dependent routing of auxiliary information into the main representation: the query-driven attention weights act as an interpretable gating mechanism that highlights auxiliary features most relevant to each SP-cVAE derived query, while mitigating the risk of overwhelming the primary model with spurious or noisy signals. Additional advantages include modality-aware feature alignment, inherent robustness to missing or degraded auxiliary inputs, and straightforward inspection of persample contribution via attention maps.

Prior to fusion, each of the three encoders was pre-trained with a multilayer perceptron regression head on the target tasks to obtain pre-trained weights; in the fusion stage, these pre-trained encoders serve as frozen-weight feature extractors. And their hidden representations are incorporated as features. To enable cross-modal interaction, the heterogeneous feature dimensions were first unified through learnable linear projection layers: the SP-cVAE features (192-dim) were projected to 128-dim, the BiG-CAE latent features (64-dim) were projected to 128-dim, and the PH-NN features (128-dim) were projected to 128-dim. This dimensional alignment ensures all modalities operate in a common embedding space, facilitating meaningful cross-attention computation. In this scheme the SP-cVAE supplies the query ($Q$) while the auxiliary branches supply keys ($K$) and values ($V$) as shown in **Fig. 2** (c), realizing the scaled dot-product attention computation:

$$\text{Attention}\left(Q,\ K,\ V\right)=\text{softmax}\left(\frac{QK^{\top}}{\sqrt{d_k}}\right)V \tag{6}$$

All pre-trained weights are frozen to protect learned representations and ensure reproducibility; a residual connection balances the primary and auxiliary pathways and prevents uncontrolled information leakage. Attended auxiliary signals are aligned, concatenated with SP-cVAE features, and passed through a lightweight fusion network. Final predictions use a residual form to prioritize SP-cVAE outputs:

$$\hat{y}_{\text{final}} = \alpha \cdot \hat{y}_{\text{SP-cVAE}} + \left(1-\alpha\right) \cdot \hat{y}_{\text{fusion}} \tag{7}$$

where $\hat{y}_{\text{fusion}}$ represents the residual produced by the fusion framework, while $\hat{y}_{\text{SP-cVAE}}$ denotes the residual output of SP-cVAE. And $\alpha$ is a learnable, softmax-normalized parameter that calibrates the auxiliary contribution without allowing it to eclipse the SP-cVAE pathway (Hyperparameters in Table S36). Training configurations of COFAP are listed in Table S37.

## Training strategy

The data partitioning strategy employed a 70:15:15 ratio for training, validation, and test sets respectively. To ensure representative distributions across all subsets, a stratified splitting approach was adopted based on the target property values. Specifically, continuous target values were discretized into quantile-based bins, and samples were allocated to each subset while preserving the overall distribution shape. This stratification ensures that the model encounters representative examples across the full property range during both training and evaluation.

Several measures were implemented to mitigate data leakage risks. First, structure-level separation was enforced such that all nine sectional projections derived from a single COF structure were assigned to the same data subset, preventing the model from exploiting structural similarities between projections of identical materials during evaluation. Second, each COF structure maintained a consistent unique identifier across all three neural network branches, ensuring that feature extraction remains consistent and prevents mixing between training and test samples. Third, all pre-trained encoders were frozen during the fusion stage with parameters fixed at values learned exclusively from the training set, eliminating information leakage through gradient backpropagation. Finally, feature standardization parameters were computed solely from the training set and applied uniformly across all subsets.

For robust performance evaluation, a 5-fold stratified cross-validation protocol was implemented following a nested design. The outer loop partitioned the complete dataset into 5-fold using stratified sampling based on binned target values, with 1-fold held out as the test set in each iteration. Within each training-validation pool, an inner loop performed an additional 5-fold stratified split to separate training and validation sets, ensuring that hyperparameter tuning and early stopping decisions were made independently of final test set evaluation. The stratification employed 5–10 bins depending on dataset size, with bin boundaries determined by quantile thresholds to ensure adequate sample representation in each stratum. All random operations were controlled using fixed seeds to guarantee reproducibility across experimental runs.

## Application of COFAP on high-throughput screening

A reproducible pipeline was designed to convert predictions into ranked and diversified candidate sets. The workflow consists of metric normalization, composite scoring, weight-sensitivity analysis, contribution-rate reporting, and aggregation of structural statistics.

**Metric normalization.** Two metrics are considered: $R$% and APS. They are normalized by min-max scaling:

$$\tilde{R}_i = \frac{R_i - \min_j R_j}{\max_j R_j - \min_j R_j}$$
$$\widetilde{\text{APS}}_i = \frac{\text{APS}_i - \min_j \text{APS}_j}{\max_j \text{APS}_j - \min_j \text{APS}_j} \tag{8}$$

**Composite scoring.** A convex combination is used to compute the composite score:

$$S_i\left(w_R, w_A\right) = w_R \tilde{R}_i + w_A \widetilde{\text{APS}}_i$$
$$w_R + w_A = 1, w_R, w_A \geq 0 \tag{9}$$

**Weight-sensitivity analysis.** To assess stability of top candidates, scores are recomputed across a grid of $(w_R, w_A) \in [0,1]\times[0,1]$. For each weight pair, the top-10 list is compared with a baseline ($w_R = w_A = 0.5$) via overlap fraction:

$$overlap\left(w_R, w_A\right) = \frac{\left|TOP\text{-}10\left(w_R, w_A\right) \cap TOP\text{-}10_{\text{baseline}}\right|}{10} \tag{10}$$

**Contribution-rate reporting.** For each candidate, metric contributions are recorded as

$$contrib_{R,i} = w_R \tilde{R}_i$$
$$contrib_{A,i} = w_A \widetilde{\mathrm{APS}}_i$$
$$rate_{R,i} = \frac{contrib_{R,i}}{contrib_{R,i} + contrib_{A,i}} \quad (11)$$
$$rate_{A,i} = \frac{contrib_{A,i}}{contrib_{R,i} + contrib_{A,i}}$$

**Feature statistics.** For each weight pair, the top-100 candidates are analyzed to extract frequency distributions of bond types, topological nets, and linkers.

All intermediate and final results are saved in readable format to ensure reproducibility.